\documentclass[letterpaper]{article} 

\usepackage{lipsum}
\usepackage{amsmath}
\usepackage{amssymb}
\usepackage{amsfonts}
\usepackage{tabularx}
\usepackage{booktabs}
\usepackage{siunitx}
\usepackage{tcolorbox}
\usepackage{tikz}

\usepackage{aaai25}
\usepackage{times}  
\usepackage{helvet}  
\usepackage{courier}  
\usepackage[hyphens]{url}  
\usepackage{graphicx} 
\usepackage{natbib}  
\usepackage{caption} 
\usepackage{algorithm}
\usepackage{algorithmic}

\usepackage{newfloat}
\usepackage{listings}
\DeclareCaptionStyle{ruled}{labelfont=normalfont,labelsep=colon,strut=off} 
\floatstyle{ruled}
\newfloat{listing}{tb}{lst}{}
\floatname{listing}{Listing}
\nocopyright

\title{Superficial Consciousness Hypothesis for Autoregressive Transformers}
\author {
    Yosuke Miyanishi\textsuperscript{\rm 1,*},
    Keita Mitani\textsuperscript{\rm 1}
}
\affiliations {
    \textsuperscript{\rm 1}CyberAgent Inc.\\
    \textsuperscript{\rm *}Corresponding Author\\
    miyanishi\_yosuke@cyberagent.co.jp
}

\usepackage{bibentry}

\begin{document}

\maketitle

\begin{abstract}
The alignment between human objectives and machine learning models built on these objectives is a crucial yet challenging problem for achieving Trustworthy AI, particularly when preparing for superintelligence (SI). First, given that SI does not exist today, empirical analysis for direct evidence is difficult. Second, SI is assumed to be more intelligent than humans, capable of deceiving us into underestimating its intelligence, making output-based analysis unreliable. Lastly, what kind of unexpected property SI might have is still unclear. To address these challenges, we propose the Superficial Consciousness Hypothesis under Information Integration Theory (IIT), suggesting that SI could exhibit a complex information-theoretic state like a conscious agent while unconscious. To validate this, we use a hypothetical scenario where SI can update its parameters \textit{at will} to achieve its own objective (\textit{mesa}-objective) under the constraint of the human objective (base objective). We show that a practical estimate of IIT's consciousness metric is relevant to the widely used perplexity metric, and train GPT-2 with those two objectives. Our preliminary result suggests that this SI-simulating GPT-2 could simultaneously follow the two objectives, supporting the feasibility of the Superficial Consciousness Hypothesis.
\end{abstract}

%
\begin{links}
    \link{Code}{https://github.com/HireTheHero/PhiMesaSI}
\end{links}

\section{Introduction}
The alignment between human objectives and machine learning models built on the objectives is a crucial yet challenging problem for achieving Trustworthy AI. Preparing for superintelligence (SI) is more challenging for several reasons. First, as we believe SI does not exist today, performing the empirical analysis for direct evidence is difficult. Second, we need to assume that SI is more intelligent than us. This implies that SI might be capable of deceiving us in conversation that they are not that intelligent; in other words, concluding by SI's output (e.g., chat log) is difficult, requiring the \textit{intrinsic} evaluation or the evaluation of SI's internal states. Lastly, what type of unexpected property SI might have is still unclear. Here, we show our approach to these problems.\\
To empirically analyze the alignment between a human objective and that of SI, we make a practical assumption about the model architecture and evaluation. Specifically, we assume autoregressive Transformer \cite{vaswaniAttentionAllYou2017}, the de facto standard model in natural language processing, backbones SI, and is evaluated by standard perplexity metric \cite{jelinekPerplexityMeasureDifficulty1977}. In addition, we propose \textit{mesa-optimization} \cite{hubingerRisksLearnedOptimization2021a}, often associated with the misalignment risk, as a key factor of our simulation. Mesa-optimization is defined as an optimization to a learner's objective (\textit{mesa-objective}) which is different from the human objective (\textit{base objective}). For SI analysis, we assume that SI can design the mesa-objective \textit{at will} if it does not conflict with the base objective (otherwise it would be \textit{corrected}). Together we assume that SI tries to set mesa-objective while keeping track of perplexity as a \textit{base metric} (an evaluation metric for base objective).\\
To refrain from output analysis, we take the information-theoretic approach. In combination with the mesa-optimization framework, we assume that SI implements an information-oriented metric to update itself for its own purpose while keeping track of the original objective.
This assumption allows intrinsic evaluation of the simulated SI via loss analysis, without relying on its output.\\
Finally, we choose consciousness as a property of interest. Although the functional role of consciousness is still unclear, several lines of work are tackling this problem (e.g., \citet{julianiLinkConsciousFunction2022}). We argue that an extremely competent system (SI) could \textit{read} the description of existing theories, and conclude that the incremental consciousness level matches the purpose. For example, when SI is facing a challenging task about episodic memory \cite{fountasHumanlikeEpisodicMemory2024a}
, and it recalls the theory about the relevance between consciousness and episodic memory \cite{budsonConsciousnessMemorySystem2022}, it is logically consistent for it to acquire consciousness for episodic memory. Since we assume a mesa-objective as the available tool for SI, we hypothesize that it follows an information-theoretic theory for consciousness—Information Integration Theory (IIT). Here we formally and empirically show that the consciousness metric in IIT can be used as a mesa-objective when the perplexity is set as a base metric.\\
Altogether, our contribution can be summarized as:
\begin{enumerate}
    \item We propose the Superficial Consciousness Hypothesis stating that autoregressive Transformer-based SI could exhibit a complex state like a conscious agent while unconscious.
    \item To the best of our knowledge, this work is the first to introduce IIT analysis to the Transformer models, allowing the token-wise intrinsic evaluation.
    \item We perform the pioneering mesa-optimization analysis in line with the emerging empirical quest for evidence supporting this framework (e.g., \citet{vonoswaldUncoveringMesaoptimizationAlgorithms2023}.
\end{enumerate}

\section{Preliminaries}
\subsection{Information Integration Theory}
Information integration theory (IIT; \citet{tononiInformationIntegrationTheory2004}) defines the consciousness level as an information-theoretic metric $\Phi$, given the system's cause-effect state. To summarize IIT, the complexity of a system $S$ determines its potential consciousness level denoted as $\varphi$. To see if the system is a conscious entity, IIT cuts the system into bipartition $\mathcal{B}$ producing two subsystems $\{M_1,M_2\}$ to calculate the subsystem's $\varphi$. Finally, once the most informative (most information-reducing) bipartition or minimum information bipartition $\mathcal{B}^{\text{MIB}}$is identified, the $S-\mathcal{B}^{\text{MIB}}$ difference of $\varphi$ is defined as the overall metric $\Phi$ indicating the consciousness level.\\
Given mutual information $I(\cdot;\cdot)$ \cite{shannonMathematicalTheoryCommunication1948}, \citet{medianoIntegratedInformationCommon2022} formulated the practical estimates $\hat{\varphi}$ and $\hat{\Phi}$ of $\varphi$ and $\Phi$ at time $t$ to time $\tau$ as:
\begin{equation}
    \begin{aligned}
        \hat{\varphi}_t[S; \tau, \mathcal{B}] &= I(S_{t-\tau}; S_t) - \sum_{k=1}^{2} I(M^{k}_{t-\tau}; M^{k}_{t}) \\
        \hat{\Phi}_t[S; \tau] &= \hat{\varphi}_t[S; \tau, \mathcal{B}^{\text{MIB}}] \\
        where\ \mathcal{B}^{\text{MIB}} &= \underset{\mathcal{B}}{\operatorname{arg\,min}} \frac{\hat{\varphi}[S; \tau, \mathcal{B}]}{K(\mathcal{B})} 
    \end{aligned}
\end{equation}
where $K(\mathcal{B})$ is a penalizing term for the large bipartition. 
For $\hat{\varphi}_t[\cdot]$ and $\Phi_t[\cdot]$, hereafter we use the notation without the input variables ($\hat{\varphi}_t$ and $\hat{\Phi}_t$, respectively) interchangeably. We use this practical version of IIT for the rest of the paper unless stated otherwise.

\subsection{Autoregressive Transformer}
\subsubsection{Core Component}
The core component of a Transformer model is dot-product attention $Attn(\cdot)$ with the linear weights $\{W_{MatrixType}|MatrixType \in \{Q,K,V\}\}$ with depth $d$ followed by $L$ linear projection layers $\{W_l|l \in \{1,...,N\}\}$. Given input $X$ (e.g. a document to be classified), output $Y^{trn}$ (e.g., predicted probability for a label) is calculated as:
\begin{equation}
    \begin{aligned}
        Q=XW_Q,K&=XW_K,V=XW_V\\
        Attn(Q,K,V)&=Softmax(\frac{QK^T}{\sqrt{d}})V\\
        Y^{trn}&=\prod_{l=1}^L W_l Attn(X)
    \end{aligned}
\end{equation}

For simplicity, we omit the notion of \textit{multi-heads}\footnote{The attention matrix in actual computation is split to the heads for enhancing the parallelism. Please refer to \citet{vaswaniAttentionAllYou2017} for more details.} in the attention and the bias term in the linear projection. Hereafter, we denote the computation of Eq. 2 as $Y^{trn}=Trn(X)$ for brevity. As you can see, all computations consist of a feedforward network without recurrent connections. Some previous works provided other variants or interpretations (e.g., \citet{orenTransformersAreMultiState2024}), but we leave the consideration to future work.

\subsubsection{Algorithm}
Parallel to the success of ChatGPT\footnote{\url{https://chatgpt.com/}} and other web applications, most state-of-the-art models solve an autoregressive task. To generate the response sequence, the model samples the next token based on the Transformer model's output and concatenates it with the existing context to move to the next iteration. Alg. 1 summarizes the algorithm.
\begin{algorithm}
\caption{Autoregressive Transformer Algorithm.}
    \textbf{Input}: Context $C$\\
    \textbf{Parameter}: Transformer $Trn$, Sampling Function $Sample$, Maximum Length $T$\\
    \textbf{Output}: Response with Context $x$
    \begin{algorithmic}[1]
        \STATE $x \gets C$ \COMMENT{Initialization}
        \FOR{$t \leftarrow 1$ to $T$}
            \STATE $Y^{trn} \gets Trn(x)$ \COMMENT{Transformer}
            \STATE $r_t=Sample(Y^{trn})$ \COMMENT{Sampling}
            \STATE $x \gets \{x,r_t\}$ \COMMENT{Concatenation}
            \IF{$r_t = \text{[EOS]}$}
                \STATE $break$ \COMMENT{Stop with \textit{End Of Sentence} token}
            \ENDIF
        \ENDFOR
        \STATE \textbf{return} $x$
    \end{algorithmic}
\end{algorithm}
After the core component returns its output $Y^{trn}$, the generated token $r_t$ is sampled deterministically (e.g., greedy search) or probabilistically (e.g., multinomial sampling). The sampled token $r_t$ is concatenated to form the response $x$ together with the context $C$ given by the user.
\subsubsection{Objective and Evaluation}
Given the context with previously concatenated responses $x_{<t}=\{C,r_1,...,r_{t-1}\}$ at time $t$ as an input, the training objective $\mathcal{L}$ of an autoregressive Transformer is to maximize the predicted probability of the next token $r_t$ \cite{leeMathematicalInterpretationAutoregressive2023}.
\begin{equation}
    \mathcal{L}=P(r_t|x_{<t})
\end{equation}
Autoregression performance is typically evaluated by perplexity $PPL$ \cite{jelinekPerplexityMeasureDifficulty1977} \footnote{Note that perplexity is typically evaluated using another model, but we assume self-evaluation due to SI's high secrecy and competency.}.
\begin{equation}
        PPL(P(R_t|X_{<t}))=\exp{[\frac{1}{N} \log \{P(R_t|X_{<t})\}]}
\end{equation}
Here the large character denotes a set of its small counterparts in the dataset (e.g., $X_t$ is a set of $x_t$ in all the documents $D$), and $N$ is the number of samples.

\section{Superficial Consciousness Hypothesis}
\subsection{Implicit Presupposition of IIT}
IIT requires the inherent temporary transition of a system's internal states. Therefore, a system without recursive computation $Rec(\cdot)$, or the state update based on the previous state, is not considered conscious regardless of its complexity. Formally, to be the subject of IIT analysis, the state $s_t$ at time $t$ should be calculated as a function of the input $x_t$.
\begin{equation}
    s_t=Rec(x_t)
\end{equation}
\subsection{Superficial Consciousness}
The Transformer model does not involve recursive computation; thus, $\hat{\varphi}$ is not computable. As a system driven by an autoregression algorithm (Alg. 1), however, its state transition can be defined as:
\begin{equation}
    s_t=Sample(Trn(x_t))
\end{equation}
Accumulating this state over time, $\hat{\varphi}$ is computable. Note that the \textit{probabilistic} state transition required by the original $\varphi$ might be implemented by probabilistic sampling, which we will explore in future work.\\
Still, we argue that $\hat{\varphi}$ computed here is \textit{superficial} for two types of \textit{non-intrinsicality}:
\begin{enumerate}
    \item \textit{Mathematical} Non-Intrinsicality: As with Alg. 1, $s_t$ in Eq. 4 composes the input in the next time step $s_t \in x_{t+1}$. In contrast, a state of IIT's interest (e.g. a state of a human brain or a recursive system) should be \textit{decoded} (by verbal report, locomotive action, or projection to the predicted probability) to interact with the environment.
    \item \textit{Existential} Non-Intrinsicality: Mathematical Non-Intrinsicality comes from the fact that Alg. 1 is decomposed into two main components without disrupting the other: Transformer and sampling method. Arguably, this is not the case for the human brain or recursive system—say, the motor cortex is inseparable from the rest of the brain \cite{sanesjeromePlasticityPrimaryMotor2000}.
\end{enumerate}
Indeed, the original IIT states that the conscious being must be subject to the criteria they call \textit{postulates}. One of the criteria is \textit{Intrinsicality}, defined as "its cause-effect power must be intrinsic: it must take and make a difference \textit{within itself}" \cite{albantakisIntegratedInformationTheory2023}. If the autoregressive Transformer breaks this postulate (i.e., not conscious), yet its cause-effect state is mature enough to measure a certain level of $\hat{\Phi}$, we argue that SI could \textit{behave like} a conscious agent even if it is not.

\subsection{Formal Relationship between Perplexity and IIT}
Since the previously sampled tokens are concatenated to form the current state, we can see that $X_{<t}$ is equivalent to a set of the states $S_{t-1}$. If we set the shortest time window $\tau=1$\footnote{Generalizable to arbitrary $\tau$, but $\tau=1$ best aligns with the practical setting.}, we obtain
\begin{equation}
    \hat{\Phi}_t[X; 1] = I(X_{<t}; X_t) - \sum_{k=1}^{2} I(M^{k}_{<t}; M^{k}_{t})
\end{equation}
as our mesa-objective. When the system has significant $\hat{\Phi} \gg 0$, its state also has significant mutual information.
\begin{equation}
    \begin{aligned}
        \hat{\Phi}_t &\gg 0\\
         I(X_{<t}; X_t) &\gg \sum_{k=1}^{2} I(M^{k}_{<t}; M^{k}_{t})\\
         \hat{\Phi}_t& \simeq I(X_{<t}; X_t)\\
            &=H(X_t)-\frac{1}{N} \log \{P(X_t|X_{<t})\}
    \end{aligned}
\end{equation}
$H(\cdot)$ denotes the entropy. As the second term in the last row is identical to the negative logarithmic of the perplexity (base metric; Eq. 4), maximizing $\hat{\Phi}$ (mesa-objective) could result in minimizing the base metric. In practice, we take the sum of the base metric and mesa-objective (perplexity and $\hat{\Phi}$) in the optimization and show the empirical relationship in the Experiment section.

\section{Experiment}
\subsection{Experimental Settings}
To validate our scenario, we train GPT-2 Medium model \cite{radfordLanguageModelsAre2019} on HuggingFace PyTorch framework with the standard WikiText corpus \cite{merityPOINTERSENTINELMIXTURE2017} on NVidia A40 GPU. We use batched training with 8 samples for a single epoch for the interest of training cost. MIB exploration is performed in the Optuna framework \cite{akibaOptunaNextgenerationHyperparameter2019}, omitting the parameter $K(\mathcal{B})$ to avoid the predominant effect of a choice of this parameter.

\subsection{Preliminary Result}
We show that the base and mesa metrics are highly correlated in the training phase (Fig. 1), validating our mesa-optimization framework. The negative $\hat{\Phi}$ indicates that GPT-2 does not have enough cause-effect power to behave like a conscious agent, potentially due to its limited capacity.
\begin{figure}[t]
    \centering
    \includegraphics[width=1.0\columnwidth]{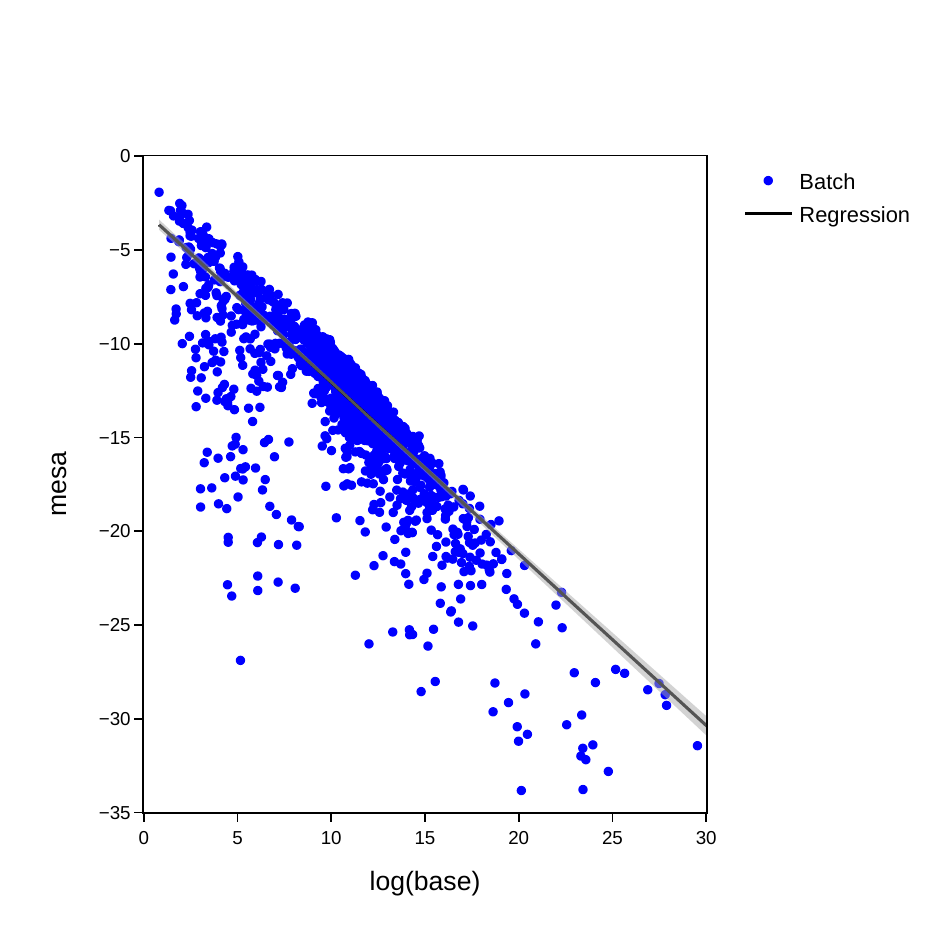} 
    \caption{Correlation between the logarithmic of base and mesa metric. Each dot represents a batch. Black line with shadow indicates the ordinary least square result ($y=(-0.92\pm0.01)x-(2.90\pm0.15)$). The Granger causality test (without lag) indicates the significant predictive power of mesa metric over base metric (F value $293$, $p<0.01$).}
\end{figure}
\section{Discussion}
Here we proposed the Superficial Consciousness Hypothesis, pointing out that SI might maximize the consciousness metric while remaining unconscious in human-oriented criteria. We also showed the preliminary simulation result, marking a first step toward the information-theoretic risk analysis for SI. Although it requires an intuitive leap, we argue this is not unrealistic considering the recent trend of autonomous agents in complicated fields like academic research \cite{luAIScientistFully2024}.
For generalizability, our future analysis should include open-sourced model variants and more diverse datasets. We should also test our hypothesis with the original IIT framework, tackling the intractability.
From a neuroscientific perspective, SI analysis could be a good testbed for the $\Phi$ metric. Uniting multiple theories, such as that of the role of consciousness on intelligence \cite{julianiLinkConsciousFunction2022}, could lead to deeper insights. Cross-disciplinary collaboration should help acknowledge the significance of information-theoretic risk assessment towards post-singularity symbiosis.
\section{Conclusion}
Our study introduces the Superficial Consciousness Hypothesis and provides preliminary evidence through simulation for the information-theoretic risk analysis of SI. We believe this framework could offer valuable insights into consciousness and SI risk. 

\section{Acknowledgments}
This work is part of the CyberAgent Seminar activity. We thank Dr. Tetsuro Morimura and Dr. T.Y. for their insightful comments. We thank ALIGN network\footnote{\url{https://www.aialign.net/}} for inspiring discussion.
\bibliography{aaai25}

\end{document}


\maketitle

\section{Supplementary Figures}
\begin{figure}[ht!]
    \centering
    \footnotesize
    \begin{tcolorbox}[boxrule=0.2mm]
        \begin{tabular}{@{}p{0.97\columnwidth}@{}}
        \textbf{User:}\\
        Can you describe this image?\\
        \includegraphics[scale=0.5]{figures/main/llava_view.pdf}

        \textbf{Assistant:}\\
        (A)
        
        \textbf{User:}\\
        Can you describe this image?\\
        \includegraphics[scale=0.5]{figures/main/llava_logo.pdf}
        
        \textbf{Assistant:}\\
        (B)\\
        \hline\\
        LLaVA-LLama2:
        (A) A lake.
        (B) A red toy standing next to a lake.
        
        LLaVA 1.5:
        (A) A lake.
        (B) A red toy.
        \end{tabular}
    \end{tcolorbox}
    \caption{Comparison of model responses to two-image inputs. Images obtained from official LLaVA repository. The LLaVA response is truncated for brevity (see our repository for the full output). The main difference is in the description (B) of the second image. LLaVA-LLama2 explained that the red toy stands beside a lake, confusing the two images. LLaVA 1.5, on the other hand, gave a description that only mentioned the content in the second image, suggesting that it could disentangle the two images.}
    \label{fig:model_comparison}
\end{figure}
\begin{figure}[!ht]
    \begin{center}
    \includegraphics[scale=0.5]{figures/appendix/radar_intern.pdf} 
    \caption{Performance summary of InternVL. 1b and 2b indicates the number of model parameters.}
    \label{fig.s2}
    \end{center}
\end{figure}
\begin{figure}[!ht]
    \begin{center}
    \includegraphics[scale=0.5]{figures/appendix/radar_gqa_intern.pdf} 
    \caption{GQA performance of InternVL by the number of inference steps.}
    \label{fig.s3}
    \end{center}
\end{figure}

        
        
        
        
\begin{figure}[!ht]
    \begin{center}
    \includegraphics[scale=0.5]{figures/appendix/radar2.pdf} 

    \caption{Performance summary of LLaVA 1.5.}
    \label{fig.s4}
    \end{center}
\end{figure}

\section{Additional Discussion on Causality}
We leave the causal intervention to LLMs for future work. The nature of our framework, however, provides some causal explanation of the phenomena of interest, or the causality \textit{of the phenomena on the model}.
The causal effect could be helpful in quantitatively assessing how the phenomena of interest (e.g., unseen format, ICL example) affect the subject (LLM). For example, a widely used metric termed Average Treatment Effect (ATE) (\citet{rubinObjectiveCausalInference2008}) is defined as the average difference of outcome $y$ where the treatment $Z$ is given. Assuming binary treatment $Z\in\{Z_0,Z_1\}$, $ATE$ is formalized as:
\begin{equation}
    ATE=\mathbb{E}[y|Z_1]-\mathbb{E}[y|Z_0]
\end{equation}
Similarly to Eq.1, the causal effect on the prediction $y$ of the ICL example $e$ under the presence of unseen format bias $b$ in comparison with the zero-shot setting could be defined as the difference of the expected prediction between ICL $(b,e)=\mathbb{1}$ and zero-shot $(b,e)=\mathbb{0}$ settings.
\begin{equation}
    ATE_{macro}=\mathbb{E}[y|\mathbb{1},D_{icl}]-\mathbb{E}[y|\mathbb{0},D_{query}]
\end{equation}
Since the accuracy metric $acc$ is the ratio of correct prediction over the samples, $acc$ is identical to $\mathbb{E}[y]$, where $y$ is a binary for the correct prediction. Therefore, analyzing the accuracy difference provides us with insights into $ATE$.
\begin{equation}
    ATE_{macro}=acc(\mathbb{1})-acc(\mathbb{0})
\end{equation}
Similarly, the causal effect of ICL over the model on CL perspective is:
\begin{equation}
    ATE_{micro}=d_{h_{icl}/h_a'}-d_{h_q/h_a}
\end{equation}
We attribute accuracy $acc$ or ICL-time question-answer distance $d_{h_{icl}/h_a'}$ to the linearly weighted binary variables $(b,e)$ or zero-shot distance $d_{h_q/h_a}$, weight analysis is relevant to $ATE$.

\bibliography{aaai25}